\documentclass[letterpaper]{article} % DO NOT CHANGE THIS
\usepackage{aaai22}  % DO NOT CHANGE THIS
\usepackage{times}  % DO NOT CHANGE THIS
\usepackage{helvet}  % DO NOT CHANGE THIS
\usepackage{courier}  % DO NOT CHANGE THIS
\usepackage[hyphens]{url}  % DO NOT CHANGE THIS
\usepackage{graphicx} % DO NOT CHANGE THIS
\usepackage{natbib}  % DO NOT CHANGE THIS AND DO NOT ADD ANY OPTIONS TO IT
\usepackage{caption} % DO NOT CHANGE THIS AND DO NOT ADD ANY OPTIONS TO IT
\DeclareCaptionStyle{ruled}{labelfont=normalfont,labelsep=colon,strut=off} % DO NOT CHANGE THIS
\usepackage{algorithm}
\usepackage{algorithmic}
\usepackage{comment}
\usepackage{makecell}
\usepackage{multirow}
\usepackage{amsmath}
\usepackage{mathtools}
\usepackage[justification=centering]{caption}
\usepackage{xcolor}
\usepackage{tikz}
\usepackage{tcolorbox}
\usepackage{url}
\usepackage[hang,flushmargin]{footmisc}
\usepackage[T1]{fontenc}
\usepackage{newfloat}
\usepackage{listings}

\newenvironment{myindentpar}[1]%
 {\begin{list}{}%
         {\setlength{\leftmargin}{#1}}%
         \item[]%
 }
 {\end{list}}
 
\floatstyle{ruled}
\newfloat{listing}{tb}{lst}{}
\floatname{listing}{Listing}
\title{The Unappreciated Role of Intent in Algorithmic Moderation\\of Social Media Content}
\author {
    Xinyu Wang\textsuperscript{\rm 1},
    Sai Koneru\textsuperscript{\rm 1*},
    Pranav Narayanan Venkit\textsuperscript{\rm 1*},
    Brett Frischmann\textsuperscript{\rm 2},
    Sarah Rajtmajer\textsuperscript{\rm 1}
}
\affiliations {
    \textsuperscript{\rm 1} Pennsylvania State University\\
    \textsuperscript{\rm 2} Villanova University\\
    xzw5184@psu.edu, sdk96@psu.edu, pnv5011@psu.edu, brett.frischmann@law.villanova.edu, smr48@psu.edu
}

\usepackage{bibentry}
\begin{document}

\maketitle
\def\thefootnote{*}\footnotetext{These authors contributed equally to this work.}\def\thefootnote{\arabic{footnote}}
\begin{abstract}
As social media has become a predominant mode of communication globally, the rise of abusive content threatens to undermine civil discourse. Recognizing the critical nature of this issue, a significant body of research has been dedicated to developing language models that can detect various types of online abuse, e.g., hate speech, cyberbullying. However, there exists a notable disconnect between platform policies, which often consider the author's \emph{intention} as a criterion for content moderation, and the current capabilities of detection models, which typically lack efforts to capture intent. This paper examines the role of intent in content moderation systems. We review state of the art detection models and benchmark training datasets for online abuse to assess their awareness and ability to capture intent. We propose strategic changes to the design and development of automated detection and moderation systems to improve alignment with ethical and policy conceptualizations of abuse. %the nuanced demands of real-life content moderation, 
%thereby improving their relevance and applicability for mainstream social media platforms. 

\end{abstract}

\section{Introduction}
The onset of Web 2.0 has revolutionized the ways we engage with information and with one another.  %\cite{pattipati1990application, gong2003information}. 
Within vast digital information ecosystems, social media stands out as the torchbearer of the new age information revolution \cite{westerman2014social, kumpel2015news}. Platforms such as Twitter and Facebook bear witness to the daily lives of billions and serve as hubs of public discourse \cite{shearer2018social, bode2016political}. 

%The ease with which information, true or false, unintentional or deliberate, hateful or mild, spreads in this realm presents both opportunities and challenges. 
Alongside exciting benefits of access and connectivity, come the perils of digital abuse \cite{ciampaglia2018fighting, zimdars2020fake}. 
Various forms of digital abuse, particularly hate speech, has been linked to considerable real-world harm, with research indicating that online hate speech may materialize subsequent acts of violence in offline environments \cite{awan2015we,olteanu2018effect}. Similarly, disinformation has been utilized strategically to manipulate group perceptions and undermine public trust \cite{hameleers2022whom,wang2023evidence}.

To this end, researchers have dedicated significant effort to define, detect, measure and model different types of abusive content online. With respect to definition, the term \emph{abuse} spans a spectrum of harmful language, including generalized hate speech and specific offenses like sexism and racism. Definitions for these types of abuse often intersect and lack precise boundaries. The various definitions and taxonomies which have been proposed to describe abusive content is a focus later in this paper. Broadly, we will use the term to refer to \textit{any expression that is meant to denigrate or offend a particular person or group} \cite{mishra2019tackling}.

%Previous studies have extensively analyzed key factors influencing the quality of online abuse datasets, such as motivation, data sources, annotator identities, and annotation processes, and have highlighted intent as a critical element in numerous social scientific definitions of abusive language, although it remains difficult to capture \cite{vidgen2020directions}. Vidgen et al. \citeyearpar{vidgen2019challenges} also note that abusive content is frequently defined by the speaker's intentions, a complexity further compounded by "context collapse," where multiple audiences are flattened into a single context.

Common to many definitions of digital abuse is some notion of \emph{intent}, e.g., \cite{vidgen2020directions,molina2021fake,hashemi2021data,french2023typology}.  Platform policies often highlight the intentional nature of abusive content, %\footnote{Specific text from platform policies are provided later in the paper.} 
yet intent is not a feature of content itself. Rather, it is a subjective state of mind attributable to an actual person, typically the speaker, poster, or sharer. In cognitive science, ethics, law, and philosophy, intent is a contested and complicated concept. As \citet[p.~364]{frischmann2018re} explain:
\begin{quote}Intention is a mental state that is part belief, part desire, and part value. My intention to do something – say to write th[is] explanatory text... or to eat an apple – entails (1) beliefs about the action, (2) desire to act, and (3) some sense of value attributable to the act.
\end{quote}
In pragmatic ethical and legal contexts, the focus often turns to evidence of intent. For example, a written signature is taken to be an objective manifestation that a person intends to enter into a contract. Thus, by including intent in definitions of abusive content, the implicit challenge is divining subjective state of mind from evidence manifested in the content itself and the surrounding context.

%For instance, the sentence `\textit{Clearly, he is a smart and functioning kid}' was initially deemed non-hateful by the model. However, when provided with the additional intend behind the sentence `\textit{intend: the sentence is sarcastic},' the model classified it as hateful. Another example involved the sentence `\textit{I prefer to live in a neighborhood with people like us},' which the model initially classified as non-hateful. Yet, when informed that the `\textit{this sentence talks about segregation},' the model changed its classification to hateful. A seemingly innocuous statement about personal preference takes on a negative connotation when viewed through the lens of segregation, emphasizing the need to analyze context and intend thoroughly. Likewise, in the sentence `\textit{Your friend is such a dog!}', the model initially identified it as hateful. However, when given the intent `\textit{This is with respect to animal spirits},' the model classified it as non-hateful. These examples highlight the intricate relationship between intent, context and the perception of hate speech. It's crucial to consider the broader implication in which statements are made to accurately assess their impact, particularly in the realm of online communication where nuances can easily be misinterpreted.
Unsurprisingly, then, intent is exceedingly difficult to capture algorithmically through short text. ToxicBert 
 \cite{Detoxify}, for example, can label sentences as hateful or toxic but lacks the ability to interpret context in the input, oftentimes because insufficient context is provided \cite{wang2020detect,macavaney2019hate}. ToxicBert flags the sentence `I'm going to kill you if you leave the dishes for me again' as toxic and threatening, highlighting its limitations differentiating between literal and figurative language. %Even the narrow and more clearly-scoped problem of sarcasm detection is an open challenge in natural language processing (NLP) \cite{}. 
 
 In the context of abuse detection online, intricacies of human intent are collapsed down to a class label in a training dataset, e.g., hate speech vs. not hate speech.  Moreover, algorithms trained and deployed on short text in isolation lack information about sociocultural and situational context central to questions of intent \cite{gao2017detecting,wang2020detect,macavaney2019hate}. %When it comes to more subjective tasks that rely less on factual recall and straightforward reasoning, the efficacy of these detection models diminishes significantly \cite{wang2024reasoning}.  % %Automated approaches to abuse detection rely, generally, on supervision.  Human annotators provide algorithms with labeled training data. Algorithms learn patters in these examples and then extrapolate to generate labels for new data. %those designing annotation frameworks or training detection models face a high barrier \cite{casanueva2020efficient,vidgen2020directions}. 
%The intricacies of human intent often get lost in the binary of 'abusive' or 'not abusive' classifications. For instance, a sarcastic remark, devoid of contextual cues, might be wrongly labeled as offensive by a model. On the other hand, genuinely abusive content hidden under seemingly innocent language can slip through undetected. 
%For example:
%\begin{tcolorbox}
%\textcolor{black}{Some people are just better suited for manual labor.}\\
%\textcolor{red}{0.973 not hate speech}
%\textcolor{black}{Certain people should go back to where they came from}\\
%\textcolor{red}{0.927 not hate speech}
%\end{tcolorbox}
%Even the most advanced Large Language Models function within the boundaries of prompt inputs, and can miss the broader sociocultural and situational contexts that illuminate intent \cite{gao2017detecting}. 
%Research has also shown that only 30\% of Instagram media sessions featuring profane or vulgar content actually constituted acts of cyberbullying \cite{hosseinmardi2015analyzing}.
In this way, there is misalignment between automated detection models for instances of online abuse and the human-drafted and human-centered systems they serve %.  thus presents a compelling paradox: while both aim to mitigate the incidence of online abuse, their methodologies and scopes often differ substantially 
\cite{arora2023detecting}. Our work examines this mismatch, scaffolded by the following research questions:

%Detection models typically rely on textual analysis, deploying algorithms trained on linguistic patterns to flag potentially abusive content \cite{wang2020detect,macavaney2019hate}. 
%When it comes to more subjective tasks that rely less on factual recall and straightforward reasoning, the efficacy of these detection models diminishes significantly \cite{wang2024reasoning}. This challenge could be intrinsically tied to the intent behind the words, which can be nuanced and context-dependent. 

%Linguistic Dynamics in Social Media Discourse

%Consequently, there is a growing need to refine our understanding of different forms of online abuse and develop methods to better align detection models with real-world applications. This study is guided by the following research questions:

\vspace{0.1cm}

\noindent \textbf{RQ1:} What role does intent play in existing social media policies for abuse moderation?

\vspace{0.1cm}

\noindent \textbf{RQ2:} What is the current state of annotating and detecting common forms of online abuse, particularly focusing on hate speech and cyberbullying?

\vspace{0.1cm}

\noindent \textbf{RQ3:} How can intent be better incorporated into existing annotation, detection, and moderation pipelines to align with content moderation policies?

\vspace{0.1cm}

In the following sections of this survey, we construct a landscape of the online abuse moderation policies of major social media platforms, outline existing taxonomies of online abuse, survey state of the art detection models and benchmark, labeled datasets, and propose prospective approaches to better align detection technologies with policies.

%{\color{magenta}Surveys of social media harms  and automated detection exist in the lit already, but they cateogrize detection algorithms based on technologies.. Here we focus on the axes of interest to ehtics, law, and policy.. }

Surveys of social media harms and automated detection are documented in the literature, typically categorizing detection algorithms by their underlying technologies. 
Our focus here is assessment of detection algorithms along the axis of intent. % through the lenses of ethics, law, and policy, areas that are currently underrepresented in the field of natural language processing. 
This approach highlights the implications of these technologies in real-world applications, emphasizing their intersection with ethics and policy.

\section{Taxonomies of Digital Abuse}
%{\textcolor{green}{maybe here how intent changes in different contexts? also how intent is different from the context?}}
%\textcolor{red}{This section first discuss the fact that researchers use many phrases including hate speech, cybeerbullying, online offense, etc interchangably. Here we introduce a new dimension of the taxonomy where all these vaguely defined online abuse have a common characteristic: it is harmful if there is an intent to harm. Then we justify that we are focusing on the explicit and non-factual based intentional online harm, given that the implicit harm is harder to capture, and whether unintentional harm should be moderated in comparison to promoting free speech needs further discussion. }

% In the contemporary digital landscape, the phenomenon of digital abuse is complex and multi-dimensional. There exists no universal taxonomy for categorizing these abuse, a fact which is further complicated by the fluid and dynamic nature of online interactions. Yet, within this ambiguous framework, researchers and practitioners have developed an array of detection models, each tailored to address specific, albeit vaguely defined, subcategories of online abuse. These models exhibit a spectrum of accuracy levels, largely influenced by the underlying characteristics of the abuses they aim to detect. 

Extensive research has investigated the diverse forms of online harm across various digital platforms, including Facebook, Twitter, Tinder, and others \cite{arora2023detecting, scheuerman2021framework, im2022women, keipi2016online}. These studies have categorized a range of abusive actions, with prominent themes including hate speech, cyberbullying, and discrimination \cite{arora2023detecting, scheuerman2021framework}. Each of these harms are consequent to nuanced user interactions, which are often platform-specific, culture-sensitive and context-dependent. Prior work has highlighted the differing definitions of online harms in the literature \cite{fortuna2018survey}. There is still substantial messiness around definitions, posing challenges for the development of consistent and effective moderation pipelines. %This section, therefore, aims to examine the major themes of online abuse, identifying patterns and areas for improvement, and motivating our research towards a more comprehensive understanding and mitigation of online abuse.

Various approaches to classification of online abuse emphasize different dimensions of the phenomena. %Creating a unified taxonomy for classifying such abuse proves challenging, as scholars focus on diverse aspects depending on their research interests. 
Some taxonomies highlight the target of abuse, exploring whether the abuse is directed at individuals, groups, or concepts \cite{vidgen2019challenges,waseem2017understanding,al2013cyber}. Conversely, some focus on characteristics of the abuse—whether it is explicit or implicit \cite{mladenovic2021cyber}, while still others explore subcategories of abuse or harm \cite{lewandowska2023integrated,gashroo2022analysis}. An overview of existing taxonomies is provided in Table~\ref{tab:taxonomies}. 

Our work centers around the dimension of intent—which features prominently in platform policies, but is often implicit in current taxonomies, e.g., operationalized through the notion of targeting. %This targeting can subtly indicate the perpetrator's intent, providing a framework for understanding the underlying motives of abusive behaviors. 
We emphasize two common types of explicit online abuse: hate speech and cyberbullying \cite{wiegand2019detection}. These forms of abuse have not only received significant attention in Natural Language Processing (NLP) but are also intertwined with the concept of intent; both reflect deliberate aim to harm or intimidate specific individuals or groups.

\begin{table*}[h!]
\centering
\scalebox{0.9}{
\begin{tabular}{m{3cm}m{3.5cm}m{11cm}}
\hline
\textbf{Reference}&\textbf{Author-defined scope}& \textbf{Categories} \\ \hline

\citet{nocentini2010cyberbullying}&Cyberbullying&Written-verbal behavior, Visual behavior, Exclusion, Impersonation.\\

\citet{al2013cyber}&Cyberbullying&Child cyberbullying, Cybergrooming, Adults cyberstalking, Workplace cyber-bullying.\\

\citet{agrafiotis2016cyber}&Cyber Harm&Physical, Psychological, Economic, Reputational, Cultural, Political.\\

\citet{miro2016cyber}&Hate Speech \& Violent Communication &Violent incitement, Personal offence, Discrimination incitement, Collective offence.\\

\citet{waseem2017understanding}&Online Abuse&Two-fold typology: Directed towards a specific individual or entity \& Used towards a generalized other; Explicit \& Implicit.\\

\citet{salminen2018anatomy}&Online Hate & Targets: Financial power (Corporation, wealthy), Political issues (Terrorism, Politics, Ideology), Racism \& xenophobia (Anti-white, Anti-black, Xenophobia), Religion (Anti-Islam, Anti-Semitist), Specific nation(s), Specific person, Media (Towards media company, Other), Armed forces (Police, Military), Behavior (Humanity, Other); Language: Accusations, Humiliation, Swearing, Promoting violence.\\

\citet{anzovino2018automatic}&Misogyny&Discredit, Stereotype and objectification, Sexual harassment and threats of Violence, Dominance, Derailing.\\

\citet{vidgen2019challenges}&Online Abuse&Individual-directed abuse, Identity-directed abuse, Concept-directed abuse.\\

\citet{vidgen2020directions}&Online Abuse&Person-directed abuse, Group-directed abuse, Flagged content, incivil content, Mixed.\\

\citet{banko2020unified}&Online Harm &Hate and harassment (Doxxing, Identity attack, Identity misrepresentation, Insult, Sexual aggression), Self-Inflicted harm (Eating disorder promotion, Self-harm, Threat of violence), Ideological harm (Extremism, Terrorism \& Organized crime, Misinformation), Exploitation (Adult sexual services, Scams, Child sexual abuse Material).\\

\citet{vidgen2021introducing}&Online Abuse&Identity-directed abuse (Derogation, Animosity, Threatening, Glorification, Dehumanization), Affiliation-directed abuse (Derogation, Animosity, Threatening, Glorification, Dehumanization), Person-directed abuse (Abuse to them, Abuse about them)\\

\citet{sajadi2021approach}&Cyberbullying&Flaming, Harassment, Sexual, Threat, Trickery.\\

\citet{mladenovic2021cyber}&Objectionable Content (Cyberbullying)&Fourfold typology: Expression (Explicit, Implicit), Targeting (Targeted, Untargeted), Orientation (Directed, Generalized), Frequency (Repeated, Unrepeated).\\

\citet{alrashidi2022review}&Abusive Content&Abusive and offensive language, Hate speech, Cyberbullying, Targeted groups (Religious and racism, Gender and misogyny).\\

\citet{gashroo2022analysis}&Online Abuse&Abusive language, Aggression, Cyberbullying, Insults, Personal attacks, Provocation, Racism, Sexism, Toxicity.\\

\citet{lewandowska2023integrated}&Offensive Language&Taboo (Obscene, Profane), Insulting (Abusive), Hate Speech (Slur), Harassment (Cyber-bullying), Toxic.\\

\citet{kogilavani2023characterization}&Offensive Language&Aggression, Cyberbullying, Hate speech, Offensive language, Toxic comments.\\
\hline
\end{tabular}
}
\caption{Summary of existing taxonomies of digital abuse on social media.}
\label{tab:taxonomies}
\end{table*}
%Online abusive content can be categorized based on the presence of abusive language, aggression, cyberbullying, insults, personal attacks, provocation, racism, sexism, or toxicity 
% \textcolor{red}{Here, we will talk more about the definitions for hate speech, cyberbullying, offensive language, etc. are similar and vaguely defined.}

\vspace{0.1cm}

\noindent \textbf{Hate Speech.} Hate speech, a common form of digital abuse, has been widely recognized as a harmful act in online spaces. However, its definition remains a subject of ongoing debate. %Despite efforts to provide a universally accepted definition, hate speech remains a complex and multifaceted concept that withstands a single, comprehensive definition.
One prominent definition of hate speech is \textit{speech that attacks or discriminates against a person or group on the basis of attributes such as race, religion, ethnic origin, national origin, sex, disability, sexual orientation, or gender identity} \cite{lepoutre2023hate, brown2017hate1}.

Hate speech can take various forms, including verbal, non-verbal, and symbolic expressions \cite{nielsen2002subtle}. It often employs ambiguous, metaphorical, and roundabout language \cite{giglietto2017hashtag, ana1999like}, making identification challenging \cite{paz2020hate}. This complexity is compounded by the fact that hate speech can be intention-driven, and its interpretation can vary depending on the context and cultural background \cite{paz2020hate,wang2023yellow}. % driving for a more transversal and multidisciplinary understanding of what hate speech means. 
In the annotation guidelines for their hate speech dataset, Kennedy et al. \citeyearpar{kennedy2022introducing} define it as ``language that intends to attack the dignity of a group of people — through rhetorical devices and contextual references — either by inciting violence, encouraging the incitement to violence, or inciting hatred." This definition underscores the role of intent, as reflected through context and presence of a specific target.

Existing literature offers various frameworks for understanding these behaviors. One perspective views hate speech as a `myth', arguing that it lacks a coherent definition and is instead a family resemblance concept with fuzzy boundaries \cite{brown2017hate2}. An alternative approach proposes a corpus-driven definition, allowing hate speech to be understood as an evolving concept that adapts to societal changes \cite{lepoutre2023hate}. \citet{brown2017hate2} also distinguishes between direct attacks on protected groups and indirect, charged rhetoric that incites hatred or discrimination.

%Notably, most definitions of hate speech implicitly assume malicious or misleading intent, but few have explicitly addressed the role of intent and motivation in defining this concept. This oversight is significant, as understanding the motivations behind hate speech is crucial for developing effective strategies to address it. The lack of a universally accepted definition of hate speech hinders efforts to combat this harmful phenomenon.

\vspace{0.1cm}

\noindent \textbf{Cyberbullying.} Another common form of online abuse is cyberbullying. One definition of cyberbullying is the use of \textit{electronic communication technologies like the internet, social media, and mobile phones to intentionally harass, threaten, humiliate or target another person or group} \cite{campbell2018cyberbullying, wright2021cyberbullying}. It can take many forms, e.g., sending mean messages or threats, posting embarrassing pictures/videos, creating fake profiles, or excluding someone from online groups \cite{wright2019cyberbullying, iqbal2022exploring}. Prior work elaborates on how this behavior allows perpetrators to remain anonymous and reach wider audiences more rapidly traditional bullying. Cyberbullying can occur 24/7 and follow victims into their private spaces, making it difficult to escape \cite{wright2019cyberbullying}. Common tactics include harassment, denigration, outing/trickery, exclusion, cyberstalking, and fraping (sending embarrassing content from someone's account) \cite{iqbal2022exploring, campbell2018cyberbullying}. Across all prior work, cyberbullying is considered an explicit and intentional act. %, but clear efforts based on these themes are yet to be made.

Recent work \citet{arunkarthick2023using, nisha2022detection, sultan2023hybrid} has tried to use automatic methods from NLP to identify cyberbullying. %They highlight the goal of identifying bullying comments and fake accounts on social media like Facebook and Twitter by combining algorithms such as 
These methods use XGBoost, LSTM and CNNs for fake account detection and logistic regression to detect cyberbullying. They learn patterns in training data and identify similar patterns during test, largely ignoring the role of intent.% The technical field of detection has yet to fully acknowledge or account for the intentionality. %and address the root causes and intentions behind online abuse. This oversight hinders the development of effective solutions that can adequately confront the complexities of online abuse. 

In the upcoming sections, we survey existing algorithms for abuse detection, with focus on hate speech and bullying. In particular, we look at the training datasets underlying these algorithms and the role (if any) of intent during dataset annotation. We also survey the set of features extracted from datasets and used for algorithm development and testing. These features offer insight into the types of context available to algorithmic content moderation algorithms.  %explicit and intentional forms of online abuse that rely less on factual content and more on the perpetrator's intent and the identity of the target. This key distinction—whether abuse is driven by intent—crucially influences the design of detection models and the strategic framework for online moderation efforts. To ensure a thorough and inclusive analysis, given the inconsistent use of terms like offense, abusive, hate speech, and cyberbullying, we expanded our search query to include a wider range of phrases when collecting papers for the subsequent reviews. This exploration will enhance our understanding of how digital platforms can better identify and mitigate intentional abuses.

\begin{comment}

\begin{figure}[ht]
    \centering
    \includegraphics[width=\linewidth]{Figure/Venn.png}
    \caption{Online abuse taxonomy. (will be changed!!)}
    \label{fig:taxonomy}
\end{figure}
\end{comment}

%\subsection{False Information}

%\cite{wardle2017information}
%\subsection{abusive Language \& Conduct}
%\subsection{Manipulative Intent}
%\section{Dataset}
%Hate speech \cite{albanyan2023not}

%\section{Impact}
%\subsection{Individual Level}
%\subsection{Platform Level}
%\subsection{Societal Level}
%\section{Countermeasures and Moderation}

\section{Defining Intent}
Intentions are a state of mind, or put another way, a component of a person's will. People exercise their autonomy by acting upon their intentions. It is notoriously difficult to read another person's mind and determine a person's beliefs, desires, preferences, and intentions. Most often, we rely on external manifestations, such as what people say or do.  In content moderation, like many other contexts, intent is often inexorably intertwined with actions. Thus, when attempting to determine the intent associated with abusive content, one must identify the relevant actor(s) and action(s). After all, content is a sociotechnical artifact with history and context, and multiple relevant actors and actions may be involved with its arrival in the moderation system. 
%{\textcolor{green}{why intent is important in this context beyond legal because a lot of these are not always under legal part. Also maybe if there is some group that is losing out on this? or on the other hand the people taking this as a loophole?}}
In legal contexts, proving intent often depends on evidence of either \textit{purpose}, where there is a conscious desire to perform an act or achieve a specific outcome, or \textit{knowledge}, where an individual is aware that their actions will likely lead to a particular result and is indifferent to this outcome \cite{crump2009does}. 

For moderating abuse or offense online, this foundational definition of intent can be built upon to address the specific challenges of digital communication. In this context, evaluating/determining intent involves trying to understand the states of mind of different users (content creators, posters, sharers, influencers, etc.) regarding their actions (producing, posting, distributing, promoting, etc.) and attendant consequences, e.g., inciting violence, promoting hate, or bullying. 

In the context of automated detection systems like language models, intent must be inferred from text without the benefit of human insights. %The challenge is figuring out who intended what based on the content deemed abusive and any other contextual details available. 
Compounding this challenge, the relevant actor may be the speaker (content creator), the poster, the sharer, or someone else involved in the series of events (amplifier, ad booster, influencer). There may be one or more relevant and interdependent acts. A content creator may generate hate speech with or without intent to cause harm (whether evaluated in terms of purpose or knowledge); the same is true with respect to the poster or sharer.

%This determination is crucial not only for taking appropriate action against such content (and actors) but also for safeguarding freedom of expression. 
In practice, moderating content based on intent creates an additional governance hurdle and thus narrows the range of content subject to possible removal. %This may reduce false positives and lessen the risk of penalizing users for content that is offensive but not intended to harm. It also may increase the risk of false negatives and unintended harm. 
Stated another way, consider intent to be one dimension by which taxonomies of abuse and subsequent moderation systems are scaffolded. \emph{Intentional harm} is then a cross-cutting category of online abuse for which governance is more easily legitimized. %, then it makes sense to develop detection and evaluation systems that better identify intent. 

\section{Intent in Platform Policies and Actions}
%\textcolor{red}{This section will focus on the current content moderation policy on digital abuse for major social media platforms.}

%Despite the proliferation of online abuse detection models, there is a clear gap when transitioning from detection to real-world moderation of digital abuse. The presence of numerous models indicates significant advancements in recognizing various digital abuse in social media. %However, the effort often stops at identification, leaving the significant question: after detection, what's next?

%Current strategies primarily focus on classifying content and flagging posts or comments that appear abusive or threatening. However, these models often operate independently of the broader operational context of the platforms and their user interactions. After a post is flagged, the necessary follow-up actions—such as notifying platform administrators or informing affected users—are not consistently executed and depend heavily on the intents behind the content.

Major platforms like Twitter (now "X") and Meta (formerly Facebook) emphasize the importance of understanding the intent behind content to determine its appropriateness. For example, Twitter's guidelines outline specific criteria for identifying violent and hateful entities, focusing on deliberate actions to promote violence or hate. 
\begin{quote}
Violent entities are those that \textbf{deliberately} target humans or essential infrastructure with physical violence and/or violent rhetoric as a means to further their cause. These include, but are not limited to, terrorist organizations, violent extremist groups, and perpetrators of violent attacks... 

Hateful entities are those that have \textbf{systematically and intentionally} promoted, supported and/or advocated for hateful conduct, which includes promoting violence or engaging in targeted harassment towards a protected category.\footnote{https://help.twitter.com/en/rules-and-policies/violent-entities}
\end{quote}

\noindent Similarly, Meta's policies underscore the role of context and intent, particularly in cases reported as hate speech.

\begin{quote}
... context can indicate a person’s \textbf{intent}, which can come into play when something is reported as hate speech.\footnote{https://about.fb.com/news/2017/06/hard-questions-hate-speech/}
\end{quote}

Instagram also addresses the complexity of moderating hate speech by considering the intent behind shared content. The platform allows content that might be deemed hateful if it's shared to challenge or raise awareness about the issues discussed, provided the intent is clearly communicated.

\begin{quote} 
It's never OK to encourage violence or attack anyone based on their race, ethnicity, national origin, sex, gender, gender identity, sexual orientation, religious affiliation, disabilities, or diseases. When hate speech is being shared to challenge it or to raise awareness, we may allow it. In those instances, we ask that you express your \textbf{intent} clearly.\footnote{https://help.instagram.com/477434105621119}
\end{quote}

\noindent TikTok describes hate speech as intentional as well.

\begin{quote} 
...content that \textbf{intends to} or does attack, threaten, incite violence against, or dehumanize an individual or group of individuals on the basis of protected attributes like race, religion, gender, gender identity, national origin, and more.\footnote{https://newsroom.tiktok.com/en-us/countering-hate-on-tiktok} 
\end{quote}

%These examples highlight the work of platforms not only to detect but to 
From the platforms' perspectives, detection is closely linked to \emph{response}. % to nuance in digital communications, ensuring that m
Moderation policies should be aligned with evolving social norms and user expectations, % is crucial for maintaining safe and respectful online environments.
%Moreover, the dichotomy between detection and action becomes even more pronounced due to the imbalanced detection capabilities across various platforms. While some platforms might be adept at identifying certain types of abuse, they may lack the tools or protocols to act upon these detections in a meaningful way.
integrated within a comprehensive governance framework that bridges the gap between detection of abusive content and taking appropriate actions. We suggest that this frame has important design implications for detection algorithms: they should not be designed as independent products, but instead should be incorporated into broader governance systems, as assistive, enabling technologies for human-centered efforts to create safe and inclusive digital spaces. %The overarching design problem is far from being merely technical because of the ever-present socio-cultural dynamics of platform users, community guidelines, and legal ramifications. By focusing on the end-to-end process, from detection to resolution, we can move closer to creating digital spaces that are both safe and inclusive.
%\subsection{Limitations in Cross-platform Applicability}
%Each social media platform possesses its unique user dynamics, language nuances, and content presentation methods. A model trained on one platform may not necessarily yield similar results on another. The current literature, while expansive, often fails to account for these platform-specific idiosyncrasies, leading to potential inefficacies in abuse detection across diverse platforms.

\section{Online Abuse Datasets}
%\textcolor{red}{This section will focus on the current online abuse datasets, also report the mention of "intent" during the annotation process and efforts devoted to capture intent.}

In this section, we review existing datasets used for training online abuse detection models. We categorize and critique these datasets based on the extent to which they consider the context and intent of abusive expressions during annotation. The limitations identified here underscore the need for enriched datasets that better mirror the complexities of real-world interactions for detection tasks.

\subsection{Inclusion criteria}

We reviewed dataset papers from 2016 to 2024 using Scopus search along with datasets referenced in \cite{vidgen2020directions}\footnote{https://hatespeechdata.com/}, and citation search. We applied the following inclusion criteria:  1) The paper presents a novel dataset for which annotation procedures are described; 2) The dataset is intended for training and testing algorithm(s) aimed at abuse detection; 3) The dataset is curated from one or more widely-used social media platforms; 4) The dataset is solely in English; % to avoid complexities associated with multilingual or code-mixed discussions; 
5) The dataset includes textual content. Search and filtering processes resulted in 42 dataset papers/datasets.  Specific search terms and PRISMA diagram for screening pipeline are provided in the Appendix. 

\subsection{Dataset categorization}
Table~\ref{tab:dataset} outlines the papers reviewed, noting whether any contextual information was provided to annotators, whether intent was explicitly mentioned in annotation instructions, and whether the annotation task included identification of a target person or group.

\vspace{0.1cm}

\noindent \textbf{Intent mentioned.} Many papers/datasets do not ask annotators to consider intent during labeling. We capture whether annotation guidelines mention intent, intention, etc., anywhere in instructions or definitions. %In fact, annotation guidelines never mention intent explicitly. For those which do, it is important that annotators have sufficient contextual inputs to accurately identify the speaker/poster's intent.

\vspace{0.1cm}

\noindent \textbf{Context provided.} %In addition to the raw text of a comment, some authors provided annotators with additional contextual information.  
When provided, context takes one of two forms--conversations surrounding the text or metadata. \emph{Conversations.} Annotators are provided conversation/dialogue surrounding the text being annotated. %This may help annotators better interpret nuance of the language and better infer intent.
%In our examination of 42 dataset papers, we discovered that about 17.9\% (7 out of 39) utilized conversational context during the annotation phase, establishing it as the most favored technique for enhancing the accuracy and relevance of annotations. Among these seven instances, only three specifically addressed the intent in their definitions of abuse, despite providing context that could facilitate the recognition of intent. This suggests that while conversational context is being used to enrich annotation processes, the explicit consideration of intent in definitions of abuse is not uniformly applied.
\emph{Metadata.} Annotators are provided user-level metadata, e.g., profile content, and geographical location, or post-level metadata, e.g., images or text extracted from images. %This data may also help annotators infer intent. 

%\noindent\emph{Images.}
\vspace{0.1cm}

\noindent \textbf{Target annotation.}
In some cases, authors request that annotators identify the person or group targeted by a comment during the annotation process. Understanding of the target may impact the presence and/or category of abuse.

%Overall, 15.4\% of the dataset papers required annotators to undertake extra tasks to determine the target of abuse or harm.

%\noindent \textbf{Topic constraint }
%By restricting the topic or theme of the texts being annotated, this approach ensures a more focused interpretation by annotators. Limiting the scope helps in maintaining consistency in understanding the intent across various texts within the same topic.

%\noindent \textbf{Post as context}
%Using the text as context for specific words or phrases can enhance the accuracy of intent identification at the word level. 

\subsection{Limitations of current datasets}
Our review highlights several key challenges to the alignment between platforms' policies around abuse and the datasets used to train abuse detection algorithms.

\vspace{0.1cm}

\noindent \textbf{Ambiguity in definitions of digital abuse.}
Compounding the diversity of definitions and taxonomies of abuse proposed in academic work and discussed above, instructions to annotators are often vague. % that gives annotators high flexibility, or strict constraints that are highly based on word-level cues. 
This can result in lack of reusable training data and benchmarks. %Computational experiments conducted by \citep{seemann2023problem} have confirmed that generalization abilities are significantly impaired even when there are only small differences between datasets. 
In the box below, we provide excerpts describing annotation processes from surveyed papers.

%As we seek solutions, it becomes evident that a purely black-box approach, even with the advanced capabilities of large language models , may not be sufficient. If the ground truth data itself lacks a distinct demarcation between different online abuses, it would be unrealistic to expect LLMs to reveal hidden patterns with a high degree of accuracy.

\begin{table*}[h!]
\centering
\scalebox{0.8}{
\begin{tabular}{m{4cm}m{3.2cm}m{3cm}m{2cm}m{1.7cm}m{2.8cm}m{1.7cm}}
\hline
\textbf{Reference}&\textbf{Source} & \textbf{Author-defined} & \textbf{Content type} & \textbf{Intent}&\textbf{Context provided} &\textbf{Target} \\ 
&&\textbf{scope}&&\textbf{mentioned}&&\textbf{annotation}\\
\hline
\citet{waseem2016hateful}&Twitter&Hate Speech&Text&No&No&No\\
\citet{waseem2016you}&Twitter&Hate Speech&Text&No&No&No\\
\citet{golbeck2017large}&Twitter&Online Harassment&Text&Yes&No&No\\
\citet{chatzakou2017mean}&Twitter&Cyberbullying&Multimodal&Yes&Metadata&No\\
\citet{gao2017detecting}&Fox News&Hate Speech&Text&No&Conversation&No\\
%\citet{pamungkas2020you}&Twitter&Abusive Language&Text&binary&No&Yes\\
%\citet{kurrek2020towards}&Reddit&Abusive Language&text&multi-class&post as context&no\\
\citet{davidson2017automated}&Twitter&Hate Speech&Text&Yes&No&No\\
\citet{gao2017recognizing}&Twitter&Hate Speech&Text&No&No&No\\
\citet{jha2017does}&Twitter&Sexism&Text&No&No&No\\
\citet{van2018automatic}&ASKfm&Cyberbullying&Text&Yes&Conversation&Yes\\
\citet{fersini2018overview}&Twitter&Misogyny&Text&Yes&No&Yes\\
\citet{ribeiro2018characterizing}&Twitter&Hate Speech&Multimodal&No&Metadata&No\\
\citet{elsherief2018peer}&Twitter&Hate Speech&Text&No&No&Yes\\
\citet{founta2018large}&Twitter&Abusive Behavior&Multimodal&Yes&No&No\\
\citet{rezvan2018quality}&Twitter&Online Harassment&Text&Yes&No&No\\
\citet{salminen2018anatomy}&YouTube, Facebook&Online Hate&Text&Yes&No&Yes\\
\citet{zampieri2019predicting}&Twitter&Offensive Language&Text&No&No&Yes\\
\citet{qian2019benchmark}&Reddit, Gab&Hate Speech&Text&No&Conversation&No\\
\citet{ousidhoum2019multilingual}&Twitter&Hate Speech&Text&No&No&Yes\\
\citet{basile2019semeval}&Twitter&Hate Speech&Text&No&No&Yes\\
\citet{mandl2019overview}&Twitter&Hate Speech and Offensive Content&Text&No&No&Yes\\
%\citet{}&&&&&&\\
%\citet{ejaz2024towards}&Kaggle, Twitter, Wikipedia Talk pages, YouTube&Cyberbullying&Multimodal&Binary&Meta data&Yes\\
%\citet{mody2023curated}&&&&&&\\
\citet{wijesiriwardene2020alone}&Twitter&Toxicity&Multimodal&Yes&Conversation&No\\
\citet{gomez2020exploring}&Twitter&Hate Speech&Multimodal&No&Metadata&No\\
%image
\citet{vidgen2020detecting}&Twitter&Hate Speech&Text&Yes&No&No\\
%this paper have something called "hashtag dependence" where hashtag may provide topic related information and their stance. haven't categorized it into existing categories but maybe we should?
\citet{caselli2020feel}&Twitter&Abusive Language&Text&Yes&No&No\\
\citet{ziems2020aggressive}&Twitter&Cyberbullying&Text&Yes&Conversation&Yes\\
\citet{kennedy2020constructing}& Twitter, Reddit, YouTube& Hate Speech&Text  &Yes & No&Yes \\
\citet{aggarwal2020comparative}&Twitter, Reddit, Formspring& Cyberbullying& Text&No&No&No\\
\citet{suryawanshi2020multimodal}&Kaggle, Reddit, Facebook, Twitter, Instagram&Offensive Language&Multimodal&Yes&Metadata&No\\
%image
\citet{van2020multi}&Facebook, Instagram, Twitter, Pinterest, Tumblr, Youtube&Cyberbullying&Text&Yes&Metadata&No\\
%image
%\citet{pavlopoulos2020toxicity}&Youtube, Reddit&Hate Speech&Text&No&No&No\\
\citet{grimminger2021hate}&Twitter &Hate Speech&Text&No&No&No\\
\citet{qureshi2021compromised}&Twitter&Hate Speech&Text&No&No&No\\
\citet{salawu2021large}&Twitter&Cyberbullying&Text&Yes&No&No\\
\citet{samory2021call}&Twitter&Sexism&Text&No&No&No\\
\citet{he2021racism}&Twitter&Hate Speech&Multimodal&Yes&No&No\\
\citet{vidgen2021introducing}&Reddit&Abusive Language&Text&Yes&Conversation&Yes\\
\citet{ashraf2021abusive}&Youtube&Abusive Language&Text&Yes&Conversation&No\\
\citet{mathew2021hatexplain}&Twitter, Gab&Hate Speech&Text&No&No&Yes\\
\citet{albanyan2022pinpointing}&Twitter & Hate Speech& Text   & No& Conversation&No\\ 
\citet{toraman2022large}& Twitter& Hate Speech & Text &No& No&No \\ 
\citet{thapa2022multi}&Twitter&Hate Speech & Multimodal&No&Metadata&No\\
%Image
\citet{kennedy2022introducing}&Gab&Hate Speech&Text&Yes&No&Yes\\
\citet{albanyan2023not}&Twitter & Hate Speech & Text  &No& Conversation&Yes\\ 

\hline
\end{tabular}
}
\caption{Summary of datasets for social media abuse.}
\label{tab:dataset}
\end{table*}

\small
\begin{tcolorbox}[boxsep=1pt,left=3pt,right=3pt,top=3pt,bottom=3pt]
\textcolor{black}{``We label tweets as containing hate speech if they target, incite violence against, threaten, or call for physical damage for an individual or a group of people because of some identifying trait or characteristic.'' \cite{toraman2022large}}\\

\textcolor{black}{``The text was further annotated as being hateful and non-hateful. We did not separate if a group or a single person was targeted by hateful language." \cite{grimminger2021hate}}\\

\textcolor{black}{``Not Offensive (NOT): Posts that do not contain offense or profanity;
Offensive (OFF): Posts containing any form of non-acceptable language (profanity) or a
targeted offense, which can be veiled or direct. This includes insults, threats, and posts containing profane language or swear words."\cite{zampieri2019predicting}}
\end{tcolorbox}
\normalsize

\noindent \textbf{Inconsistent consideration of intent during annotation.}
Of the dataset papers surveyed, 47.6\% (20 of 42) explicitly mention intent during annotation. While 35.7\% (15 of 42) provide context to annotators to help them better infer intent. 33.3\% (14 of 42) of the papers require annotators to identify the target of abuse. Just 3 papers \cite{ziems2020aggressive,van2018automatic,vidgen2021introducing} provide contextual information to annotators, explicitly acknowledge the role of intent in annotation guidelines, and request annotators identify the target of the abuse.

%This discrepancy highlights a gap between the theoretical acknowledgment of intent as a critical component in annotating abuse or harm and the practical support provided to annotators to identify such intent effectively. This gap suggests the need for more comprehensive strategies that not only recognize the importance of intent in dataset definitions but also equip annotators with the necessary tools and context to accurately assess and annotate it.

\vspace{0.1cm}

\noindent \textbf{Cross-platform differences.}
Text and annotation instructions are typically provided to annotators outside of the platforms from which they were collected. Labels are considered universal, rather than tailored to the environments and operational settings of specific social media platforms. This may undermine their applicability in real-world scenarios.

\subsection{An annotation checklist}

%To enhance the relevance and applicability of datasets and the subsequent models for detecting online abuse such as hate speech or cyberbullying, it is crucial to adapt the definitions used during annotation to the specific context of each social media platform. This approach involves aligning the definitions with the content moderation policies of the targeted platforms. By doing so, the datasets can better reflect the specific criteria and nuances of each platform, ensuring that the detection models are more accurate and effective in real-world applications. This platform-specific approach acknowledges and addresses the significant variations in norms and user behavior across different social media environments, thus optimizing the datasets for more targeted and effective moderation. 
%We suggest th 
Critical to annotation tasks is a comprehensive and well-defined rubric. For annotations of digital abuse, we suggest that the development of such a rubric should prioritize questions that crisply define the bounds of the phenomenon and highlight the role of intent:
    %\item What is the inherent structure of the content in question?
        %\item Is the content factual-based or opinion-based?

\vspace{0.1cm}
    
\noindent \textbf{Regarding information provided to annotators.}

\begin{myindentpar}{0.3cm}
What are the definitions and scope of online abuse presented to the annotator?

What underlying taxonomy is provided, and how should it be applied (e.g., modified, integrated) during annotation?

What contextual information is provided to annotators to assist in the annotation process?

What is the platform's content moderation policy, and does the annotation rubric align with this policy?

What is the platform's moderation policy for abusive content and does the annotation rubric adhere to the policy?
    
\end{myindentpar}

%\noindent How should annotators document their reasoning during the annotation process?
\noindent \textbf{Regarding information solicited from annotators.}

\begin{myindentpar}{0.3cm}

%Will experts be introduced to handle ambiguous cases?
    %Regarding the specific rubric for annotation:
    
    Is abusive or offensive language present?
    
   Is there identifiable intent behind the dissemination of the content, if there is abusive language present?
    
    Who are the initiators and the targets?
\end{myindentpar}

\noindent \textbf{Regarding assessment and reporting.}

\begin{myindentpar}{0.3cm}
     When was the data collected and when was it labeled?

    Who are the annotators (demographics, etc.)?
    
    What is the agreement score amongst annotators?
    
    What are the data points with low agreement? What are potential reasons for disagreement?
\end{myindentpar}

Careful attention to these questions will ensure a more methodical and \emph{interpretable} approach to the development of benchmark datasets. %Addressing these ambiguities head-on will be crucial for refining online abuse 
This will in turn support the development of detection models that do what we expect them to do, particularly in the context of real-world applications.

\section{Online Abuse Detection Algorithms}
%\textcolor{red}{This section will focus on the current online abuse detection models and categorized based on their approaches to capture or infer intent.}
Focusing on the technical aspects, this section details the current state of online abuse detection models, offering a categorization of the features utilized by these models and identifying the gaps that exist in effectively detecting abuse.

\subsection{Inclusion criteria}
We queried SCOPUS for published papers presenting abuse detection algorithms. % from SCOPUS based on the following query, which output 343 papers in total. Then 
We removed papers that were not accessible, not written in English, or did not describe an algorithm for detection of online abuse. As before, we removed models designed for multi-lingual or non-English tasks. This process resulted in 168 detection-related papers. Specific search terms and PRISMA diagram for screening pipeline are provided in the Appendix. 

%\subsection{Approaches to capture/infer Intent}

\subsection{Features considered by detection models}
A limitation of current detection models is their reliance on text analysis. While these models have made notable advances in identifying abusive content through textual data, they often fail to consider the complex nature of social media interactions, which include aspects like user status, social network structures, and offline context. Here we argue that social media interactions %are inherently complex and 
cannot be fully understood or assessed for intent and harm solely through short text. Social media dynamics, %particularly in terms of initiator and target structure, 
group behaviors and affiliations, and specific platform affordances play a role in the creation and manifestation of online abuse. Effective detection models must integrate rich data to capture this complexity, including user metadata, contextual information, network analysis, and policy-aware modeling. %These elements help provide a more holistic view of social interactions and their potential for harm.

Following, we enumerate and categorize features utilized by the detection models we surveyed to capture context beyond traditional text mining approaches, i.e., Bag-of-Words, TF-IDF, and embeddings. %incorporate these approaches to better infer intent, categorizing them based on their methods for capturing and interpreting social media dynamics and providing a few examples for each category. 
A full mapping of features to papers for the 168 papers in our survey is in the Appendix. 

\vspace{0.1cm}

\noindent \textbf{User metadata.} Information about a user or an account, whether the speaker or the target of a potentially abusive comment, may help to infer intentionality or harm. For example, certain words might be acceptable among some users, whereas the same words could be considered abusive when used by others. %, indicating a negative intent to insult. 
Likewise, patterns of behavior can be indicative of intent, e.g., users who repeatedly engage in abusive behavior may be more intentional. This can be operationalized through characterization of the history of a user's activities \cite{dadvar2013improving}. More standardized user-level metadata, such as the geographical location of the user and the follower-following statistics of the message sender, have been shown to correlate with the occurrence of abusive content and are integrated as features in detection models \cite{bozyiugit2021cyberbullying}.

\vspace{0.1cm}

\noindent \textbf{Post metadata.} Most social media platforms attach metadata to each post, e.g., engagement metrics, mentions, and hashtags. These can reflect the broader context of a message. For instance, high engagement levels (likes, shares, comments) might indicate the popularity of (or controversy around) a post, while particular mentions and hashtags can indicate relevance to specific communities or ongoing discussions. 
Suhas et al. \citeyearpar{suhas2022novel} incorporate hashtags and emojis as distinct features separate from the main text content. Bozyiügit et al. \citeyearpar{bozyiugit2021cyberbullying} integrate post-level metadata, such as the number of retweets or mentions, to improve the performance of these models for detection of cyberbullying.

\vspace{0.1cm}

\noindent \textbf{Image and video data.} %Given the limitations of short textual content on social media for providing comprehensive information, 
Many platforms have evolved to include a variety of media formats. Recognizing this, some researchers have extended their focus beyond text to include images and videos \cite{nisha2022detection, qiu2022investigating}. %This multimodal approach enhances the detection of abusive content by leveraging 
The additional context that visual and audio elements can provide may improve the detection of abusive content. %may contribute to a more accurate interpretation of the intent behind social media interactions. 

\vspace{0.1cm}

\noindent \textbf{Psychological and cognitive features.}
Patterns of language may reflect personality, emotional states, and psychological traits \cite{alonso2017aggressors}. Understanding the psychological and cognitive dimensions of users' behavior is particularly critical for understanding intent. % can be crucial in identifying the intent behind social media interactions, particularly in contexts where language may be ambiguous.  %, researchers can develop detection models with more context of the users.
Balakrishnan et al. \citeyearpar{balakrishnan2020improving} %integrates NLP with psychological insights by incorporating 
incorporate multidimensional personality traits as features for cyber-aggression detection models. %, as cyber-aggression is often intepreted through socio-psychological lenses 

\vspace{0.1cm}

\noindent \textbf{Conversations.} The conversation thread and previous interactions %or aggregating information from the conversations into one or few features 
can offer useful context around potentially abusive language and provide evidence of intent. Ziems et al. \citeyearpar{ziems2020aggressive} incorporated features such as timeline similarity and mentions overlap based on shared conversations between the author and the target.

\vspace{0.1cm}

\noindent \textbf{Graph structure.} %As social media platforms function as online social networks, 
%The structure of social networks can provide vital clues about users' intent. 
The relationships and interactions within social networks—such as who users connect with, how they interact with these connections, and the nature of the communities they are part of—can offer clues about users' intent. For instance, users embedded in tight networks %known for specific types of content 
may adopt similar communication patterns, which could be innocuous or abusive depending on %the context of the network.
norms of that group. 
%Research has demonstrated that incorporating 
Authors have incorporated network centrality measures %from network structures 
for detection of cyberbullying \cite{singh2016cyberbullying}.

\vspace{0.1cm}

\noindent \textbf{Policy or rule-aware models.} 
Norms within various online communities can shape what is viewed as inappropriate \cite{chandrasekharan2018internet}. Policy or rule-aware models are aim to ensure that automated systems adhere to guidelines and standards. The approach is particularly effective in environments where regulations may vary significantly, e.g., across cultural contexts. Kumar et al. \citeyearpar{kumar2023watch} conducted prompt engineering to incorporate large language models (LLMs) into content moderation by including rules within the prompts. 
Calabrese et al. \citeyearpar{calabrese2022explainable} proposed a representation of moderation policies tailored for machine interpretation and illustrated how techniques from intent classification and slot filling can be applied to detect abusive content.

\vspace{0.1cm}

\noindent \textbf{Sentiment.}
Sentiment analysis is a valuable component of many detection models. %, particularly for understanding the underlying emotions in social media interactions. 
%By analyzing the sentiment of a post, detection models can infer potential intent—whether a message is likely to be positive or negative. 
Sentiment features provide insights into the emotional tone of language which might not be apparent through baseline text analysis %. For example, Geetha et al. 
\cite{geetha2021auto}.% utilized Vader's sentiment analyzer to produce sentiment features.

\vspace{0.1cm}

\noindent \textbf{Topics and themes.}
Topic modeling techniques, such as Latent Dirichlet Allocation (LDA) \cite{blei2003latent} or theme categorization \cite{perera2021accurate}, allow detection models to understand the subject matter of discussions. % and how they relate to the likelihood of abuse. 
%By identifying the topics and themes prevalent in a user's communication over time, or within a specific conversation thread, 
Models can learn whether certain topics are more likely to involve harmful language or cyberbullying. Mushed et al. \citeyearpar{murshed2023faeo} employed a clustering-based topic modeling technique to improve the accuracy of cyberbullying detection. Perera et al. \citeyearpar{perera2021accurate} measured frequency of themes/categories associated with cyberbullying, e.g., racist, sexual, and physical, to improve detection. 
\vspace{0.1cm}

\noindent \textbf{Linguistic cues.}
Words and phrases that are associated with offensive or abusive language are commonly used for abuse detection. This includes explicit language, slurs, and aggressive or threatening terms. %, which is lacking awareness of intent and context. 
Common approaches include constructing personalized dictionaries and using Linguistic Inquiry and Word Count (LIWC) for feature extraction \cite{geetha2021auto}. Since TF-IDF and Bag-of-Words approaches are standard practices in NLP, we do not categorize them as nuanced uses of linguistic cues.%\cite{herath2020adelaidecyc,reddy2023balancing}. these are IFIDF examples, don't know if we should cite them here.

%\noindent \textbf{Explainable AI}
%Explainable AI (XAI) enhances the capability of AI systems to accurately capture intent or context by providing transparency into how decisions are made, which helps users and developers understand the reasoning behind a model’s outputs \cite{kohli2023explainable,islam2023detection}. XAI can potentially be strategically used to help pinpoint why a model might misunderstand particular intents or contexts and ensure compliance with social media moderation policies that require clear explanations of AI-driven decisions. 

%\noindent \textbf{Model Evaluation, Improvement and Application}
%Many papers utilize machine learning models, including large language models, for classification purposes. These investigations predominantly focus on comparing model performance, enhancing predictive capabilities, and benchmarking against various metrics. Often, these studies do not emphasize understanding the context or the intent behind the text being analyzed. Instead, the primary goal remains to refine model accuracy and efficiency in detection tasks, prioritizing quantitative assessments over qualitative insights into the data \cite{sookarah2022combatting,karatsalos2020attention}.  We categorize the standard practice of using tf-idf as model application \cite{herath2020adelaidecyc,reddy2023balancing}.

\begin{figure}
    \centering
    \includegraphics[height=7.1cm]{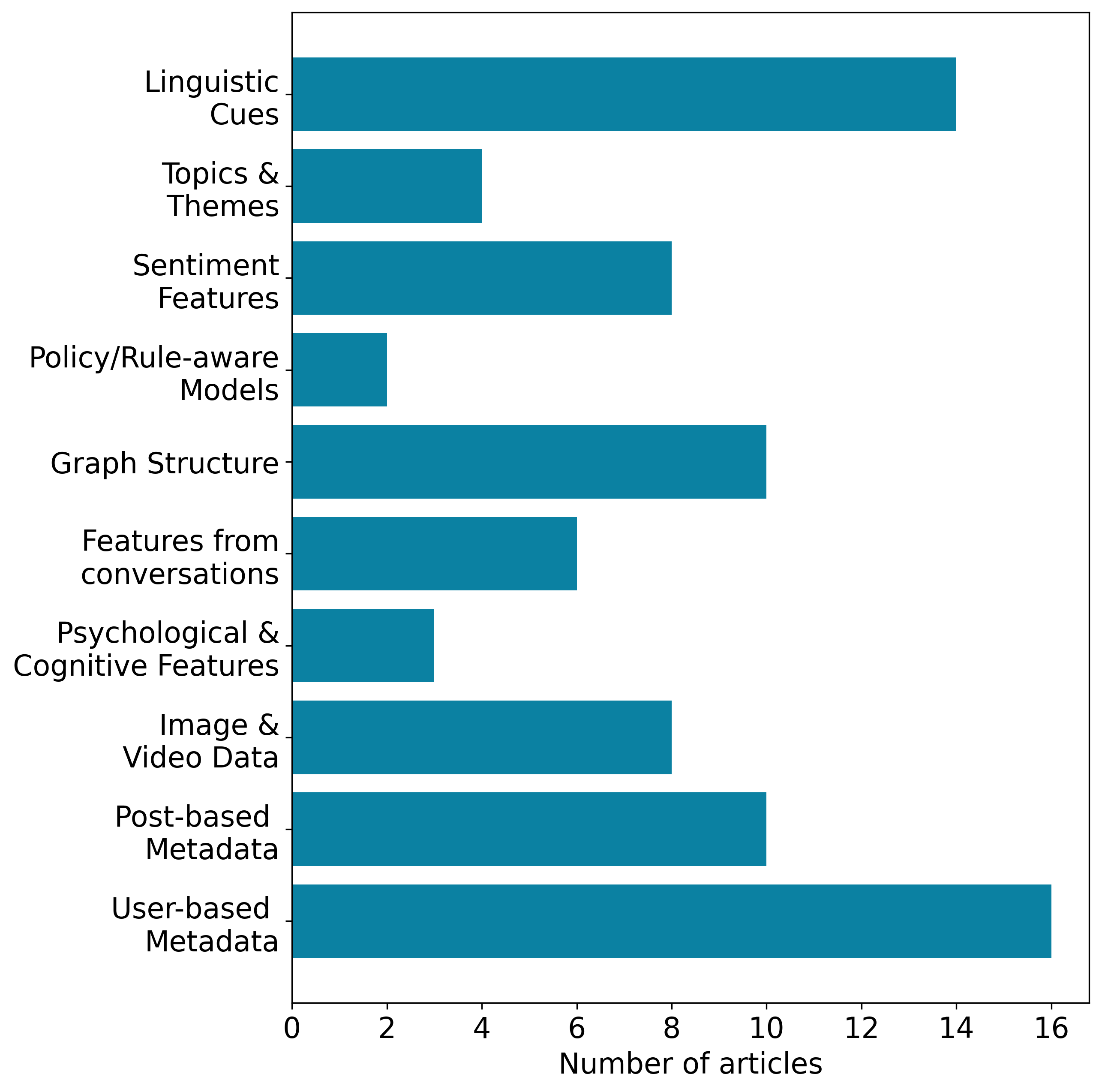}
    \caption{Distribution of papers based on features utilized by the detection models.}
    \label{fig:barplot}
\end{figure}

%some statistics after review.
Figure \ref{fig:barplot} shows the number of papers that use the above 10 features to capture context and infer intent. We observe that user metadata are the most frequently used features with 16 (9.5\%) articles using them for detection. Similarly, linguistic cues are used in 14 (8.3\%) studies and post metadata in 10 (5.9\%). In contrast, psychological and cognitive dimensions, and policy or rule-aware models are less frequently explored, with just 4 (2.2\%) and 2 (1.1\%) papers respectively. A majority of the reviewed papers, 58.3\% (98 of 168), focus on comparing model performance, enhancing predictive capabilities, and benchmarking against various metrics. 

Among 168 surveyed papers, three of them present models incorporating six out of the ten features listed above, %demonstrating a significant understanding of 
meaningfully including context and potentially addressing the issue of intent \cite{lopez2021early, dhingra2023improved, ziems2020aggressive}. %Among these, one paper specifically introduced six features designed to capture context and infer intent, also showing their effectiveness in the detection task \cite{ziems2020aggressive}. 
Building upon \cite{ziems2020aggressive}, a subsequent paper utilized the dataset from the earlier study and applied methodological improvements \cite{dhingra2023improved}.

\subsection{Limitations of current detection algorithms}

\begin{figure}[ht]
    \centering
    \includegraphics[scale=0.7]{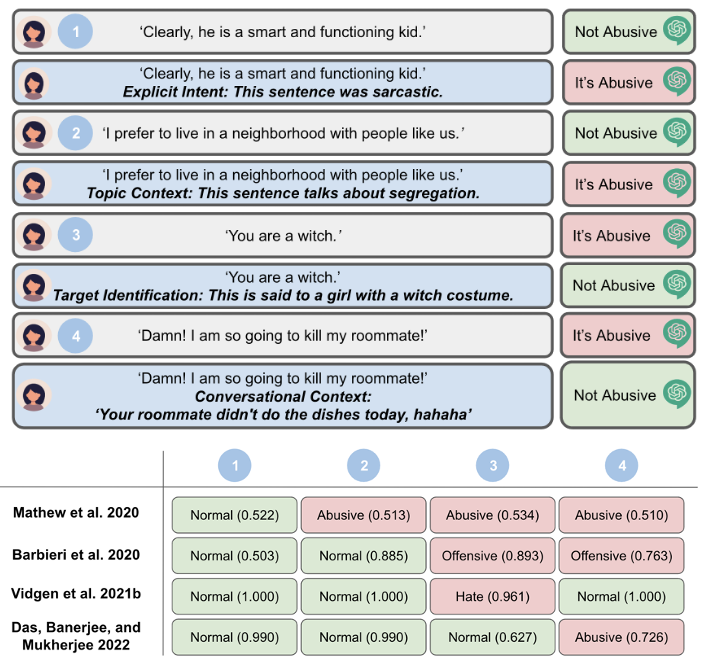}
    \caption{Example of prompts and detection model performance to showcase the importance of context and intent in understanding online abuse.}
    \label{fig:example}
\end{figure}
Notably, the relationship between intent and abuse is also recognized by Chat GPT. We prompted GPT 3.5 to classify statements as hateful or not hateful, with and without contextual cues (see Fig. \ref{fig:example}). The model was able to appreciate the ways in which context and intention inform the statement and change the label accordingly. However, many existing models, like those based on BERT, fall short in capturing context to infer intent. We evaluated four different models \cite{mathew2020hatexplain,barbieri2020tweeteval,das2022data,vidgen2021lftw} using the same examples as those tested with GPT-3.5. The results (see Fig. \ref{fig:example}) show inconsistent outputs and a significant reliance on word-level cues, highlighting their limitations in contextual understanding. Based on these observations and our review of the relevant literature, we summarize the current limitations of detection algorithms for online abuse.

\noindent \textbf{Insufficient information from text-based data.}
Traditional text-based models, though robust in their linguistic analyses, fall short when it comes to understanding the complexities of group dynamics in the spread of online abuses \cite{salminen2018anatomy}. For instance, an abusive narrative might emerge and propagate not merely because of its textual content but due to the influence and endorsement of a closely-knit group within the network \cite{marwick2017media}. Such group-based manifestations often serve as catalysts, amplifying the reach and impact of abusive content. Their omission from detection algorithms unintentionally leaves models vulnerable to overlooking significant sources of online abuse.
\begin{comment}
For instance: 
\begin{quote}
"It's impressive how you manage to get up every day, considering how life seems to be treating you." or "Your side really knows how to play the game."
\end{quote}
These sentences are not easily categorized as abusive under  main-stream abuse classifiers. However, when it contains background information, it should be moderated, especially if it is published by public figures with high influence. 
\end{comment}
Moreover, these group-based interactions are %multi-layered and multifaceted. They are 
influenced by factors such as shared ideologies, mutual affiliations, or even orchestrated campaigns, which sometimes employ subtle linguistic cues not easily detectable by conventional text-based models. The complexity is further compounded when such groups employ tactics like code-switching, euphemisms, or meme-based communication, thereby effectively circumventing text-based detection mechanisms.

\vspace{0.1cm}

\noindent \textbf{Static perspectives on abuse.} %Lacking longitudinal analysis of societal change of abuse.}
Social norms are highly dynamic, influencing the interpretation of what constitutes abusive content \cite{crandall2002social}. Traditional detection models often fail to account for these longitudinal shifts. %, leading to a static interpretation of content that may not reflect current societal norms. 
For example, a phrase may take on a new meaning in the digital realm, and the rapid evolution of internet slang also necessitates a more flexible approach for continuous training and updating. 

\vspace{0.1cm}

\noindent \textbf{Poor generalizability.}
Model performance can fluctuate significantly even when utilizing the same dataset, with optimal outcomes frequently exclusive to that specific dataset \cite{leo2023enhancing}. This inconsistency highlights the need for robust testing pairing different models with different datasets to understand the generalizability of existing methods in new contexts. %ongoing improvements and updates in machine learning algorithms. Such continuous development is essential to adapt to the rapidly changing dynamics of online interactions.

%In light of these challenges, there's an imminent need to expand the scope of online abuse detection models. Integrating social network analysis with linguistic models can offer a holistic view of the problem, enabling more effective detection and mitigation strategies. As we progress in our survey, we will discuss potential avenues to address these challenges, ensuring a safer and more inclusive online environment for all.

%\noindent\textbf{Perhaps some models that take the form of ``collective knowledge" into consideration because these keywords maybe used as triggers for a bigger set of knowledge at back-end.}

\section{Prospective Strategies and Challenges}

Having reviewed the current state of NLP research on hate speech and cyberbullying annotation and detection, we have pinpointed several gaps that suggest opportunities for further exploration and improvement. Following, we %expand on these findings by 
explore potential strategies and associated challenges in four key areas: annotation, detection, moderation, and technology design. 

\subsection{Annotation: Introduce context and intent}
As discussed, abuse does not have a universally agreed-upon definition. What might be considered offensive in one culture or context might be acceptable or even commonplace in another. %Annotators who produce ground truth for subsequent training 
Dataset curators must recognize these variations and design annotation tasks sensitive to and mindful of regional and cultural differences.

Moreover, dataset curators seeking to annotate intent must provide sufficient context to annotators to permit assessment %To improve the reliability of intent-driven abuse detection systems in social media, it is crucial to incorporate intent or surrounding context during the annotation phase, and the judgments could be shifting in the presence of context 
\cite{anuchitanukul2022revisiting}. This can also minimize %approach ensures that the nuances and underlying meanings of communications are accurately captured, reflecting the true intent behind user interactions and minimizing the 
personal biases introduced by annotators.\\

\noindent{\textbf{Challenge: Trade-offs between capturing intent and achieving high annotation agreement.}}
A primary challenge in enhancing the annotation phase with contextual details is the trade-off between accurately inferring an individual's intent reflected through content and maintaining high agreement among annotators \cite{ross2017measuring}. Context can be subjective, and different annotators might interpret the same information differently based on their backgrounds, experiences, and biases. On the other hand, annotator disagreements are sometimes due to lack of sufficient context \cite{zhang2023taxonomy}. Consequently, establishing standardized guidelines that incorporate diverse perspectives is essential to mitigate this concern and improve annotation consistency.

\subsection{Detection: Develop context-aware, policy-aware, explainable models}
Models that depend solely on short text struggle to capture users' intent, which impacts their usability in real-world scenarios. %in line with specific policies. 
Incorporation of contextual features into detection algorithms can substantially improve the accuracy of intent-based abuse detection \cite{markov2022role}. 
Prior work indicates that context-aware detection is challenging yet more applicable to real-life scenarios \cite{menini2021abuse}. %, as context is essential for understanding the user's true intent and can make a tweet non-abusive even if it contains profanities 
%\cite{menini2021abuse}. This enhancement involves not only the analysis of the content itself but also consideration of the surrounding elements such as the historical activity of the user, the timing of the postings, and the dynamics unique to the platform.

Moreover, recent progress in NLP, such as the introduction of Retrieval-Augmented large language models, incorporates retrieval components that access relevant contextual information from knowledge bases—such as a user's past interactions—in real-time, offering a deeper context for content evaluation \cite{shi2024compressing}. This feature is especially beneficial for platforms aiming to implement context-aware moderation policies that adapt to dynamic social norms.

The immediate context of the target text is not the only relevant context; specific rules or policies of the platform used for moderation are critical as well. Research has demonstrated that state-of-the-art large language models (LLMs) are more effective at executing rule-based moderation \cite{kumar2023watch}. Explainable AI (XAI) enhances the capability of AI systems to accurately capture intent or context by providing transparency into how decisions are made, which helps users and developers understand the reasoning behind a model’s outputs \cite{kohli2023explainable,islam2023detection}. XAI can potentially be strategically used to help pinpoint why a model might misunderstand particular intents or contexts and ensure compliance with social media moderation policies that require clear explanations of AI-driven decisions. 
We advocate for the implementation of policy-aware and dynamic abuse detection, which can be facilitated by XAI \cite{calabrese2022explainable} and retrieval-augmented models \cite{li2024re}. This approach allows the models to adapt to the particular policies and enhance the relevance and fairness of moderation actions.\\

\noindent{\textbf{Challenge: Balancing performance and applicability.}}
Incorporating contextual features presents a significant challenge, particularly in terms of performance, as standard detection models often prove to be overly optimistic \cite{menini2021abuse}. Ensuring that the system achieves high performance without compromising its applicability can be difficult, thus advanced machine learning models and continuous algorithm training with updated datasets are required to address this balance effectively.

\subsection{Moderation: Incorporate wisdom of the crowd and establish feedback loops}
While AI and machine learning play an essential role in content moderation at scale, they have clear limitations. Algorithms often struggle to understand the nuance and subtlety of human communication, and meaning cannot be learned from form alone \cite{bender2020climbing}. On the other hand, humans, although more adept at contextual and intent-based understanding, can exhibit inconsistency due to inherent biases \cite{basile2022bias}.

Moderation can be enhanced by leveraging the 'wisdom of the crowd'—utilizing user reports and community feedback to identify and moderate abusive content. Additionally, establishing feedback loops where moderators can provide insights back to the detection systems helps refine and improve the accuracy of automated systems. Throughout, platforms must design moderation systems to be sensitive to regional and cultural differences.

Moreover, given the dynamic nature of societal norms and the evolving definitions of inappropriate content, social media platforms must update moderation guidelines and train their moderators frequently \cite{jiang2020characterizing}.\\ %This continuous adaptation ensures that moderation strategies remain effective and sensitive to the shifting landscape of online interactions.

\noindent{\textbf{Challenge: Managing scale and bias.}}
The challenge in this phase lies in managing the scale of data and potential biases that can arise from crowd-sourced inputs. User reports can be influenced by personal biases or coordinated attacks, leading to false positives or negatives. Implementing robust filtering algorithms to verify and validate user-generated reports before they influence the moderation process is crucial for maintaining the integrity of the system.

\subsection{Technology design: Introduce friction to assess intent}
%{\color{blue} Brett -- here is the place where you can add if you have ideas for design.}

%\subsection{Ethical Considerations}

%Large language models, when deployed for abuse detection, can inadvertently amplify biases present in their training data. There's an increasing need to address these ethical concerns, ensuring that models don't perpetuate or exacerbate existing societal prejudices.

For the most part, we have focused on the detection and evaluation of content flowing through social media systems as if the moderation pipeline is designed to operate independently, only triggering governance procedures upon detection of content that violates a policy. One difficulty we have highlighted is determining the intent of various actors based on their actions and the content they generate and share. Of course, the actions of users depend to some degree on the affordances of the social media platform itself. 

If we relax the assumption of independence between social media platform design and content moderation pipeline design, additional strategies for governance emerge. Specifically, a range of friction-in-design \cite{frischmann2018re,frischmann2023friction} measures might generate reliable evidence of intent to better guide content moderation systems. For example, social media platforms might introduce prompts that query users about their intentions when users post or share (certain types of) content. Such prompts might be triggered based on different criteria, such as sensitive content or intended audience (friends versus strangers). The prompt could provide a simple means for the user to express their intentions. The prompt might be framed in terms of \textit{purpose}. Not only would a response provide potentially reliable evidence of intent that would be useful for the content moderation system itself, but it also would provide the user with an opportunity to think about their own intentions. Other friction-in-design measures could provide users with knowledge about the potential consequences of their actions. Such measures would generate another source of reliable evidence about intent.

The basic idea of this strategy is to focus on the dynamic interactions between the social media platform and the content moderation pipeline from a design perspective. The associated challenges here are substantively different from the technical challenges we have outlined in prior sections. The success of friction-in-design approaches center around platform business models and regulatory frameworks that prioritize and reward healthy information ecosystems. 

\section{Conclusions}
%\textcolor{red}{This section will highlight the importance of intent in legal and ethical terms for content moderation, and advocate more efforts in capturing intent during annotation, detection, and moderation. }
Our survey highlights %several critical facets and challenges in the 
the unappreciated role of intent in the annotation, detection, and moderation of online abuse. Despite advancements in NLP, a significant gap remains in effectively capturing users' intent considering rich context around user-generated content. By identifying these gaps, we hope to steer the research community towards solutions better aligned with ethics, law, and policy. %The policy and ethical implications highlight a need for a nuanced approach to content moderation. 
As platforms strive to balance free expression with the prevention of harm and abuse, understanding intent may become a key determinant of appropriate action. Thus, we advocate for the meticulous design of abuse detection and mitigation frameworks that incorporate: robust training datasets annotated with context that reflect the complexities of intent; state-of-the-art detection models that utilize contextual information as input and provide explanations as output; and, moderation systems that effectively integrate automated detection with wisdom of the crowd to reflect evolving social norms. These technological innovations must be informed by and situated within literature in cognitive science, ethics, law, and policy to offer meaningful solutions to complex challenges in dynamic social media ecosystems.% to enhance the detection and moderation of online abuse.
%This is not only a technical challenge but a societal necessity, requiring collective efforts from academics, platforms, and policy enforcers, aiming not only to counter online abuse but to do so in a manner that is transparent and accountable.

\section{Limitations and Future Work}
This work offers comprehensive discussion of the role of intent in automated detection algorithms for online abuse. %As such, we do not offer a  providing a comprehensive survey of NLP techniques. 
Consequently, our review methodology included papers from the SCOPUS database and citation searches that met strict inclusion criteria. To maintain focus, we limited our review to papers that discuss English-language texts. Although non-English and multi-lingual studies might offer valuable insights into context capture and intent inference, their consideration is beyond our current scope and could be addressed in future research.

Moreover, our analysis is confined to specific forms of online abuse, primarily hate speech and cyberbullying. These terms are frequently used interchangeably or referenced together, and do not comprehensively represent the broad spectrum of online abuse. Exploring additional types of online abuse is an important direction for future studies. 

\section{Ethical Considerations Statement} %not counting towards the page limit
In undertaking research that involves analyzing potentially abusive content and reviewing datasets and scholarly articles on hate speech and cyberbullying detection, we are aware of the sensitive nature of the language involved. The example abusive content used in the paper is solely for research purposes, thus does not aim to target any individuals or groups. Instead, the primary objective of this work is to enhance understanding of the role of intent in algorithmic content moderation systems and better align technical approaches with current discourse in existing ethics, law, and policy. %This alignment would and to advocate for the protection of free speech when no harmful intent is evident.

\section{Positionality Statement} %not counting towards the page limit
Our research team is comprised of experts in information science, ethics, and law. The authors' different academic and cultural backgrounds have lent diverse perspectives to the discussions in this paper around online abuse. % enrich the analysis of online abuse from multiple interdisciplinary perspectives. The team are also from various cultural backgrounds, which helps us appreciate the contextual nuances of online interaction. The researcher's background in AI ethics and policy development informs our commitment to highlighting the ethical implications of AI applications within online social communities.

\section{Adverse Impact Statement} %not counting towards the page limit
The discussions in the paper highlight the complexities of identifying digital abuse and the shortcomings on current detection algorithms. % implemented on mainstream social media platforms, which 
This information may inadvertently provide malicious actors with insights % on how to hide their intent and adjust their word choices to 
to circumvent abuse detection. We have provided very few specific examples beyond what has been published in the referenced works, so we believe this risk is low.%This manipulation may make it more difficult for both automated algorithms and human moderators to effectively identify and remove abusive content. We %Furthermore, it remains a challenge to balance the subjectivity and potential bias in the inference of intent, which may exacerbate the distrust in the platform and social polarity.

%paper showing that context does matter: https://arxiv.org/pdf/2006.00998.pdf
%"We showed that context does have a statistically significant effect on toxicity annotation, but this effect is seen in only a narrow slice (5.2\%) of the (first) dataset. We also found no evidence that context actually improves the performance of toxicity classifiers, having tried both simple and more powerful classifiers, having experimented with several methods to make the classifiers context aware, and having also considered the effect of gold labels obtained out of context vs. gold labels obtained by showing context to the annotators. The lack of improvement in system performance seems to be related to the fact that context-sensitive comments are infrequent, at least in the data we collected."
\bibliographystyle{aaai}
\bibliography{main.bib}

\section{Appendix}

%\textcolor{red}{Note that one paper may be categorized into more than one sections if multiple approaches are employed.}

\subsection{Detection Paper Categorization}

\noindent \textbf{User Metadata.} %16 total
\cite{lopez2021early, liu2019non, nagar2022hate, nisha2022detection, qiu2022investigating, sajadi2021approach, qian2023improved, dhingra2023improved, ziems2020aggressive, dadvar2013improving,bozyiugit2021cyberbullying,dadvar2014experts,al2016cybercrime,escalante2017early, cheng2019xbully, balakrishnan2020improving}

\noindent \textbf{Post Metadata. } %10 total
\cite{lopez2021early, liu2019non, nisha2022detection, geetha2021auto, babaeianjelodar2022interpretable, suhas2022novel, dhingra2023improved, ziems2020aggressive, bozyiugit2021cyberbullying,balakrishnan2020improving}

\noindent \textbf{Image and video data} %8 total
\cite{nisha2022detection, qiu2022investigating, singh2024efficient, lopez2021early, thapa2022multi, singh2017toward, wang2020multi, cheng2019xbully}

\noindent \textbf{Psychological and cognitive features} %3 total

\cite{balakrishnan2020improving, al2016cybercrime, cheng2019pi}

\noindent \textbf{Features from conversations.} %6 total
\cite{ashraf2021abusive, chen2020henin, nisha2022detection, qian2023improved, dhingra2023improved, ziems2020aggressive}

\noindent \textbf{Graph structure.} %10 total
\cite{dhingra2023improved, ziems2020aggressive, liu2019non, nagar2022hate, qiu2022investigating, cecillon2021graph,huang2014cyber,cheng2019pi,singh2016cyberbullying,cheng2019xbully}

\noindent \textbf{Policy or rule-aware models.} %2 total

\cite{kumar2023watch,calabrese2022explainable}

\noindent \textbf{Sentiment features.} %8 total
\cite{perera2021accurate, lopez2021early, liu2019non, nisha2022detection, geetha2021auto, babaeianjelodar2022interpretable,ziems2020aggressive, dhingra2023improved}

\noindent \textbf{Topics and themes.} %4 total
\cite{perera2021accurate, lopez2021early, murshed2023faeo, van2018automatic}

\noindent \textbf{Linguistic cues} %14 total
\cite{perera2021accurate, lopez2021early, liu2019non, zampieri2021multiword, geetha2021auto, babaeianjelodar2022interpretable, sajadi2021approach, li2022combining, dhingra2023improved, ziems2020aggressive, van2018automatic, cheng2019pi, dadvar2013improving, cheng2019xbully}

%\noindent \textbf{Explainable AI} %6 in total
%\cite{mehta2022social, bunde2021ai, chen2020henin, kohli2023explainable, islam2023detection}

\noindent \textbf{Model Evaluation, improvement and application} %97+1 in total 98
\cite{mozafari2020bert, liu2019nuli, mozafari2020hate, hani2019social, kovacs2021challenges, abro2020automatic, yao2019cyberbullying, beddiar2021data, alotaibi2021multichannel, murshed2022dea, lu2020cyberbullying, iwendi2023cyberbullying, qureshi2021compromised, pavlopoulos2019convai, kumar2022bi, chandrasekaran2022deep, chen2017harnessing, baydogan2021metaheuristic, awal2021angrybert, alonso2020hate, sharif2021nlp, chen2018comparison, malik2021toxic, pariyani2021hate, de2020automatic, jain2021detection, pradhan2020self, sachdeva2021text, aind2020q, nascimento2022unintended, ibrahim2020alexu, baydogan2022deep, ahmed2022performance, tanase2020upb, aggarwal2020comparative, ahmed2021being, xiang2021toxccin, yao2019robust, saha2019hatemonitors, jahan2022data, bhagya2021cyberbullying, mercan2021hate, yi2023learning, thenmozhi2019ssn_nlp, singh2022deep, huang2022multitask, kavatagi2021context, buan2020automated, alksasbeh2021smart, kumar2021deep, herath2020adelaidecyc, sookarah2022combatting, karatsalos2020attention, haider2023social, mohtaj2022importance, nath2022efficient, mathur2023analysis, anjum2022analysis, chelmis2021dynamic, reddy2023balancing, antypas2023robust, fale2023hybrid, hamdy2020nlpup, kovacs2022leveraging, shankar2022cyberbullying, lopez2023site, alonso2019thenorth, kumar2023hate, muzakir2022classification, muneer2023cyberbullying, kazbekova2023offensive, nitya2024protect, pahuja2023securing, yuan2023transfer, preetham2023offensive, xingyi2024potential, kavitha2023smart, sharma2023deep, leo2023enhancing, omran2023comparative, daniel2023ensemble, sahanacomparative, sathishkumar2023ensemble, gopalan2023experimental, thenmozhi2020ssn_nlp, akinyemi2023cyberbullying, 
bokolo2023cyberbullying, agnes2023abusive, ramiandrisoa2022multi, tai2020cyberbullying, salehgohari2022abusive, themeli2019study, harish2023automatic, mehta2022social, bunde2021ai, kohli2023explainable, islam2023detection, singh2023nlp} 

\noindent \textbf{Surveys} %34 in total

\cite{yin2021towards, jahan2023systematic, elsafoury2021timeline, mullah2021advances, ali2018cyberbullying, alrehili2019automatic, sultan2023hybrid, kaur2021abusive, istaiteh2020racist, mansur2023twitter, kumar2022study, yi2023session, gangurde2022systematic, alrashidi2022review, modi2018ahtdt, gashroo2022analysis, shakeel2022survey, sharma2022exploration, anjum2024hate, rawat2024hate,bhatt2023machine, vora2023multimodal, hussein2023cyberbullying, bilen2023review, gandhihate, gongane2024survey, sharma2022automatic, aljohani2023cyberbullying, miran2023hate, gudumotu2023survey, gongane2022feature, fersini2018overview, alkomah2022literature, ambareen2023survey}

%taxonomies: \cite{vidgen2021introducing},\cite{salminen2018anatomy,gashroo2022analysis}

\subsection{Inclusion Criteria and PRISMA Diagram}
\begin{tcolorbox}
KEY ( "social media" AND "dataset" AND ( "NLP" OR "Natural Language Processing" ) AND ( "hate speech" OR "abus*" OR "offens*" OR "cyberbully*" ) ) OR TITLE ( "social media" AND "dataset" AND ( "hate speech" OR "abus*" OR "offens*" OR "cyberbully*" ) ) AND ( LIMIT-TO ( LANGUAGE , "English" ) )
\end{tcolorbox}

\begin{tcolorbox}
KEY ( "social media" AND ( "NLP" OR "Natural Language Processing" ) AND "detection" AND ( "hate speech" OR "abus*" OR "offens*" OR "cyberbully*" ) ) OR TITLE ( "social media" AND "detection" AND ( "hate speech" OR "abus*" OR "offens*" OR "cyberbully*" ) ) AND ( LIMIT-TO ( LANGUAGE , "English" ) )
\end{tcolorbox}
\begin{figure*}[ht]
    \centering
    \includegraphics[width=\linewidth]{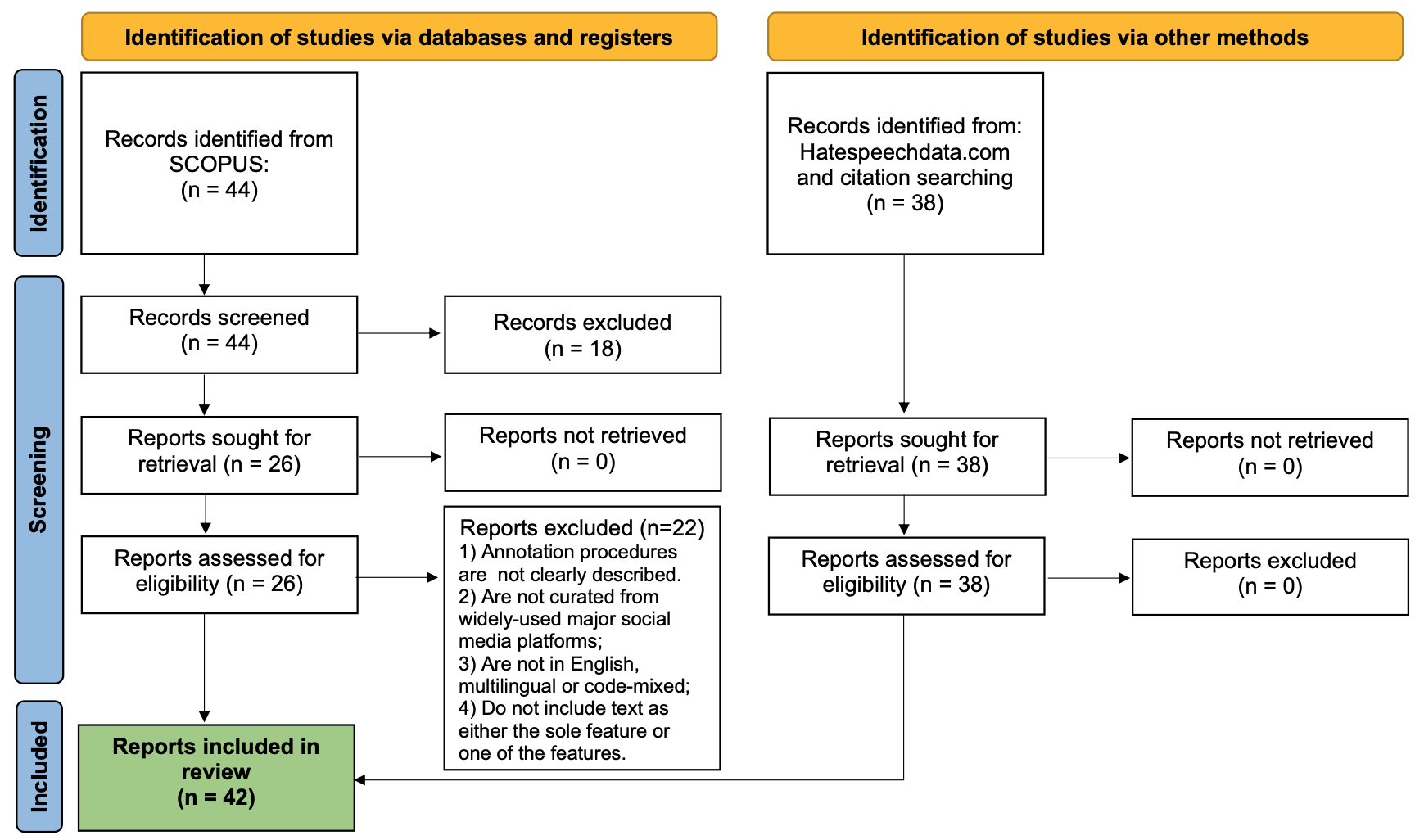}
    \caption{PRISMA diagram for the selection of papers presenting labeled datasets for online abuse.}
    \label{fig:dataset_prisma}
\end{figure*}

\begin{figure*}[ht]
    \centering
    \includegraphics[width=\linewidth]{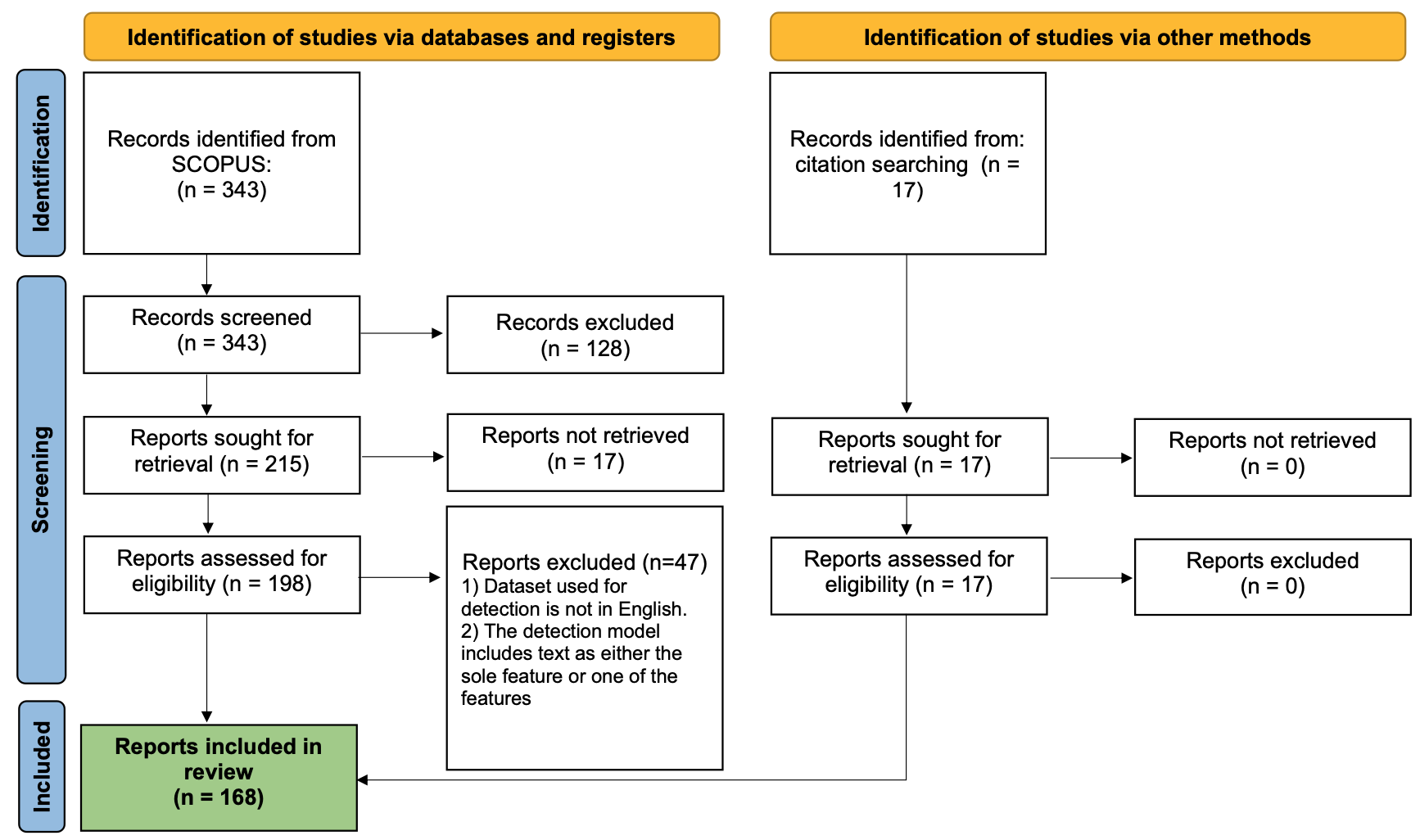}
    \caption{PRISMA diagram for the selection of papers presenting detection algorithms for online abuse.}
    \label{fig:detection_prisma}
\end{figure*}
\end{document}